\documentclass[conference]{IEEEtran}
\IEEEoverridecommandlockouts

\usepackage{cite}
\usepackage{amsmath,amssymb,amsfonts}
\usepackage{graphicx}
\usepackage{textcomp}
\usepackage{xcolor}
\usepackage{booktabs}
\usepackage{multirow}
\usepackage{array}
\usepackage{url}
\usepackage{hyperref}
\usepackage{mathtools}
\usepackage{enumitem}
\usepackage{float}
\usepackage{algorithm}
\usepackage{algorithmic}
\usepackage{threeparttable}
\usepackage{stfloats}
\usepackage{balance}
\usepackage{longtable}

\def\BibTeX{{\rm B\kern-.05em{\sc i\kern-.025em b}\kern-.08em
    T\kern-.1667em\lower.7ex\hbox{E}\kern-.125emX}}
\begin{document}

\title{Interpretable  Deepfake Detection in Videos via Explicit Forensic Features and Temporal Modeling
}

\author{\IEEEauthorblockN{ Chahira Benhama}
\IEEEauthorblockA{\textit{LARIVIA laboratory} \\
\textit{Dep. of Computer science and Eng.}\\
\textit{University of Quebec in Outaouais}\\
\textit{Gatineau, QC,  Canada}\\
\textit{Email: benc28@uqo.ca.} }\\
\and
\IEEEauthorblockN{Mohand Saïd Allili}
\IEEEauthorblockA{\textit{LARIVIA laboratory} \\
\textit{Dep. of Computer science and Eng.}\\
\textit{University of Quebec in Outaouais}\\
\textit{Gatineau, QC,  Canada}\\
\textit{Email: mohandsaid.allili@uqo.ca.} }\\
\and
\IEEEauthorblockN{  Assia Hamadene}
\IEEEauthorblockA{\textit{LARIVIA laboratory and } \\
\textit{Mixed Research Unit INRS-UQO}\\
\textit{University of Quebec in Outaouais}\\
\textit{Gatineau, QC, Canada} \\
\textit{Email: assia.hamadene@uqo.ca}} \\
}

\maketitle

\begin{abstract}
Deepfake detection in videos remains challenging, as manipulated content may appear visually consistent at the frame level while exhibiting subtle temporal inconsistencies. This paper introduces an interpretable deepfake detection framework that models spatially and temporally coherent facial features in video sequences. Unlike end-to-end deep models relying on implicit representations, the proposed approach explicitly encodes physically grounded forensic cues, enabling transparent analysis and improved multi-dataset generalization. The pipeline transforms videos into identity-consistent facial trajectories, segments them into fixed-length temporal windows, and represents each frame using 68 structured descriptors spanning four complementary domains: photometric, textural, geometric, and compression-based features. These descriptors provide a compact multi-domain representation of manipulation artifacts and are processed by a Long Short-Term Memory (LSTM) network to capture temporal dependencies and subtle irregularities. Evaluation on four benchmark datasets, FaceForensics++, Celeb-DF v2, a curated subset of the DeepFake Detection Challenge (DFDC), and DeeperForensics, yields strong and consistent F1-scores of 98.0\%, 91.0\%, 97.6\%, and 96.2\%, respectively. The approach also demonstrated a good cross-dataset generalization, providing a robust and interpretable solution for video deepfake detection.
\end{abstract}


\section{Introduction}
The rapid evolution of generative models, particularly GANs and diffusion models, has significantly increased the realism and accessibility of deepfake videos, posing major challenges for multimedia forensics \cite{Pei2026BenchmarkSurvey}. Modern pipelines generate highly realistic faces even under real-world conditions such as compression, resizing, and motion, making frame-level detection increasingly unreliable \cite{Wang2020CNNDetect}. The deepfake landscape now spans diverse forgery types and post-processing conditions, exposing the limitations of methods tailored to narrow benchmarks. Consequently, recent research has shifted toward modeling temporal consistency, improving cross-domain generalization, and enhancing explainability for trustworthy forensic analysis \cite{Rossler2019FaceForensics, Wang2024ReliabilitySurvey}.

Recent studies show that while spatial artifacts can be minimized at the frame level, deepfake generation often fails to preserve temporal consistency in videos, leading to subtle irregularities in motion, appearance transitions, and identity coherence across frames~\cite{TI2Net2023,Ramanaharan2025}. These inconsistencies, unstable facial dynamics, illumination shifts, geometric distortions, or evolving compression traces,  make temporal modeling essential for robust detection~\cite{Zhang2024Spatiotemporal}. However, many existing approaches rely on end-to-end deep representations that lack interpretability and may overfit to dataset-specific artifacts, limiting forensic reliability and cross-dataset generalization~\cite{Yan2025PlugPlay,Kim2025TemporalFrequency}. Furthermore, temporal models commonly process independently sampled frames without preserving identity-consistent facial trajectories, introducing background variations and unrelated visual content into the learned representation~\cite{Ramanaharan2025,Deng2026EfficientNetTransformer,Han2025FacialComponent}. These limitations highlight a key weakness: \textit{manipulated frames may appear realistic in isolation, while their inconsistencies become apparent only when forensic cues are analyzed coherently over time}.

Motivated by these limitations, we propose a contextual deepfake detection framework that models spatiotemporal consistency through structured and interpretable representations. Instead of relying on fully end-to-end processing, the pipeline is decomposed into principled stages: faces are first detected and tracked into identity-consistent trajectories, segmented into fixed-length temporal windows, and encoded at each frame using 68 semantically grounded features spanning \textit{photometric}, \textit{textural}, \textit{geometric}, and \textit{compression} characteristics. These feature categories are deliberately selected to capture complementary traces of face manipulation. They also encompass the principal visual inconsistencies commonly associated with face synthesis, blending, and post-processing \cite{Ramanaharan2025,Malik2022Survey}. \textit{Photometric} descriptors characterize inconsistencies in color, illumination, and local intensity statistics; \textit{textural} features capture abnormal skin patterns and fine-scale artifacts resulting from generative reconstruction \cite{Hamadene2026}; \textit{geometric} descriptors encode distortions or instabilities in facial structure and landmark relationships \cite{Belguesmia2025}; and \textit{compression} features reveal discrepancies caused by repeated encoding, resampling, and manipulation-specific post-processing. 

Temporal dependencies among frame-level representations are then modeled using an LSTM to identify persistent inconsistencies, abrupt variations, and unnatural temporal evolution \cite{Yang2024ChTransformer}. The originality of the proposed approach lies in combining explicit forensic descriptors with temporal sequence learning over identity-consistent trajectories. This enables the model to capture coherent temporal patterns overlooked by frame-based methods while reducing overfitting to dataset-specific visual cues. Moreover, the use of physically meaningful features improves interpretability by linking detection decisions to observable forensic inconsistencies. Our contributions are summarized as follows:

\begin{itemize}
\item A complete pipeline that transforms raw videos into homogeneous, identity-preserving face sequences, ensuring temporal coherence and avoiding frame-level ambiguities. The method introduces a compact set of 68 interpretable forensic features across photometric, textural, geometric, and compression cues, providing a structured and physically grounded representation of manipulation traces, with improved robustness and interpretability.

\item A unified spatiotemporal framework that encodes each identity-consistent track as a fixed $T \times 68$ matrix, standardizing variable-length videos while preserving temporal dynamics. An LSTM model is used capture subtle inconsistencies beyond frame-level analysis. Extensive evaluation on FaceForensics++, Celeb-DF v2, DFDC, and DeeperForensics demonstrates strong and consistent performance across diverse conditions. Cross-dataset evaluation on ForgeryNet and WildDeepfake further demonstrates the framework ability to generalize to  real-world deepfake distributions.
\end{itemize}

The rest of this paper is organized as follows: Section \ref{litterature_review} presents a literature review. Section \ref{framwork} introduces the proposed framework. Section \ref{experiments} presents the experimental results. We end the paper with a conclusion and future work perspectives.

\section{Recent Literature and Positioning}
\label{litterature_review}
Recent work has shifted deepfake video detection toward more realistic and comprehensive evaluation settings. Recent benchmarking efforts unify tasks, datasets, and metrics, highlighting the need for robust evaluation under rapidly evolving generative models \cite{Pei2026BenchmarkSurvey}. Similarly, Celeb-DF++ emphasizes cross-dataset testing across diverse forgery types, compression levels, and visual qualities \cite{Li2025CelebDFPP}. These studies demonstrate that performance on a single benchmark is insufficient, as detection difficulty varies considerably across generation pipelines and post-processing conditions. In parallel, several approaches explicitly address cross-dataset generalization \cite{Guo2025TemporalForgery,Yermakov2025Generalizes}. Overall, these works establish generalization and multi-benchmark evaluation as central requirements for reliable deepfake detection.

A second major trend is the shift from static spatial analysis toward temporal and contextual modeling. Recent approaches exploit global and local temporal consistency, spatiotemporal patterns, and frequency-based cues to capture artifacts that remain invisible at the frame level \cite{Amin2024TemporalCoherence,Yan2025PlugPlay,Kim2025TemporalFrequency}. These studies reinforce the view that videos should be treated as dynamic signals rather than collections of independent frames. Beyond temporal cues, contextual modeling has gained importance, with methods incorporating multi-face and scene-level relationships or combining multi-view and temporal inconsistencies for detection and localization \cite{Hu2025HICOM,Zhang2025MVIM}. These works highlight the importance of relational and contextual information in realistic scenarios. Finally, interpretability is emerging as a key requirement, with recent work combining temporal modeling with more transparent decision mechanisms \cite{Yoshii2025FairInterpretable}.

In this context, the present work positions itself at the intersection of temporal modeling and explainability, proposing a sequence-based framework that operates on explicit, semantically meaningful features. Rather than replacing end-to-end architectures, it offers a complementary and more interpretable paradigm for deepfake video forensics.

\section{Overall Framework}
\label{framwork}
Let a video be represented as a sequence $\mathcal{V} = \{I_t\}_{t=1}^{T_v}$, where $I_t \in \mathbb{R}^{H \times W \times 3}$ is the $t$-th RGB frame and $T_v$ is the number of extracted frames after temporal harmonization \cite{Huang2020}. If the original frame rate is $f_{\text{src}}$ and the target frame rate is $f_{\text{tar}}=25$, a standard temporal resampling map is
$
t_k = \left\lfloor \frac{k\,f_{\text{src}}}{f_{\text{tar}}} \right\rfloor, k=1,\ldots,T_v
$.
This step ensures that all videos contribute to the sequence model under the same temporal resolution. For every frame $I_t$, we apply the Retina Face detector \cite{Deng2020} to produce a set of candidate faces: $\mathcal{D}_t = \left\{ d_t^{(1)}, d_t^{(2)}, \ldots, d_t^{(n_t)} \right\}$, where each detection is represented by a vector $
d_t^{(i)} = \left(x_t^{(i)}, y_t^{(i)}, w_t^{(i)}, h_t^{(i)}, s_t^{(i)}\right)^T
$.

Here, $(x_t^{(i)}, y_t^{(i)})$ denotes the upper-left box corner, $(w_t^{(i)}, h_t^{(i)})$ its width and height, and $s_t^{(i)}$ the detection confidence. Note that to preserve facial geometry, each detected face crop is resized to $224 \times 224$ using an aspect-ratio-preserving strategy: the crop is first scaled so that its longest side fits within $224$ pixels, then padded with black pixels (value 0) on the remaining sides to fill the target resolution. 

\subsection{Face Tracking Formulation}
To ensure temporal consistency, detected faces are linked across frames into identity-preserving tracks, each represented at time $t$ by a state vector:
\begin{equation}
\mathbf{x}_t = (u_t, v_t, a_t, h_t, \dot{u}_t, \dot{v}_t, \dot{a}_t, \dot{h}_t)^T,
\end{equation}
where $u_t$ and $v_t$ denote the center coordinates of the bounding box, $a_t$ is the aspect ratio, $h_t$ is the box height, and $\dot{u}_t$, $\dot{v}_t$, $\dot{a}_t$, and $\dot{h}_t$ are their temporal velocities. The track state is predicted at the next frame using a linear motion model \cite{Farahi2020}: $\mathbf{x}_{t|t-1} = \mathbf{F}\mathbf{x}_{t-1|t-1}$, where $\mathbf{F}$ is the state transition matrix. The uncertainty of the prediction is represented by the covariance matrix $\mathbf{P}_{t|t-1}$. A face detection is represented by the measurement vector $\mathbf{z}_t = (u_t, v_t, a_t, h_t)^T$, 
which contains only the observed bounding-box location and size parameters. The observation model is defined by matrix $\mathbf{H}$, which maps the predicted state to the measured variables. After associating a detection with a track, the state is updated as
\begin{equation}
\mathbf{x}_{t|t} = \mathbf{x}_{t|t-1} + \mathbf{K}_t(\mathbf{z}_t - \mathbf{H}\mathbf{x}_{t|t-1}),
\end{equation}
where $\mathbf{K}_t$ is the Kalman gain, controlling how much the prediction is corrected by the new observation. $\mathbf{P}$ is the state covariance matrix, $\mathbf{F}$ the motion model, $\mathbf{H}$ the observation matrix, and $\mathbf{K}_t$ the update gain. This allows robust and smooth temporal tracking of facial regions across  the video.

\subsection{Identity Association Cost}

For each face detection $i$ and active track $k$, the identity assignment cost is defined as:
\begin{equation}
C(i,k) = \lambda_{\text{m}} C_{\text{m}}(i,k) + \lambda_{\text{a}} C_{\text{a}}(i,k),
\end{equation}
where $\lambda_{\text{m}} + \lambda_{\text{a}} = 1$. The term $C_{\text{m}}(i,k)$ is the motion consistency term defined by the Mahalanobis distance $C_{\text{m}}(i,k)
=
\sqrt{
\left(\mathbf{z}_t^{(i)}-\hat{\mathbf{z}}_t^{(k)}\right)^T
\mathbf{S}_t^{-1}
\left(\mathbf{z}_t^{(i)}-\hat{\mathbf{z}}_t^{(k)}\right)
}$, where $\mathbf{S}_t$ is the covariance matrix of the prediction error associated with the Kalman filter, capturing the uncertainty of the predicted state at time $t$, and $\hat{\mathbf{z}}_t^{(k)} = \mathbf{H}\mathbf{x}^{(k)}_{t|t-1}$ is the predicted observation. The term $C_{\text{a}}(i,k)$ is the appearance consistency term given by the cosine distance between the current embedding and the track embedding:
\begin{equation}
C_{\text{a}}(i,k)
=
1-
\frac{
\mathbf{r}^{(i)}_t \cdot \bar{\mathbf{r}}^{(k)}
}{
\|\mathbf{r}^{(i)}_t\|_2 \|\bar{\mathbf{r}}^{(k)}\|_2
}.
\end{equation}
The optimal assignment is then obtained via the Hungarian algorithm:
$
\pi^\star = \arg\min_{\pi} \sum_{(i,k)\in\pi} C(i,k)
$.

A valid temporal sequence is constructed by enforcing strict consistency constraints for reliable spatiotemporal analysis. Each sequence contains 20 consecutive frames from the same tracked face, preserving identity coherence. Temporal order is maintained to capture facial dynamics, and all frames are normalized to $224 \times 224$ for consistent feature extraction. Finally, multiple overlapping sequences are generated using a sliding-window strategy with a fixed length $T=20$ and stride $s=10$. This approach increases the number of training samples while preserving temporal continuity.

\section{Explicit Per-Frame Feature Families}
\begin{figure}
    \centering
    \includegraphics[width=\linewidth]{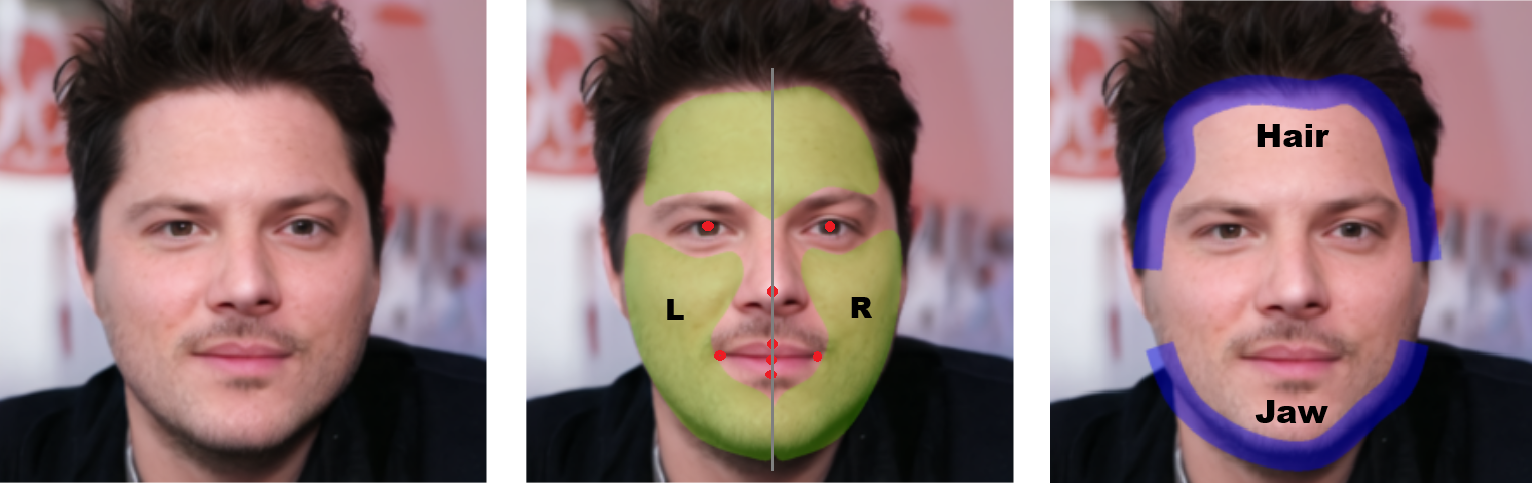}
    \caption{Facial analyzed area: (Left) Original input face image, (Middle) Segmented facial region (green overlay) where color and texture features are extracted. Automatically detected landmarks (red dots)  used to compute geometric features. The midline divides the face into left (L) and right (R) halves, enabling quantitative assessment of facial symmetry. (Right) is the analyzed regions covering hair-skin contact and jaws.}
    \label{fig:face_regions}
\end{figure}

Each cropped face is represented as a feature vector $\mathbf{x}_t \in \mathbb{R}^{68}$, comprising descriptors from four complementary families, 18 photometric, 24 textural, 14 geometric, and 12 compression-related,  as detailed in Table~\ref{tab:all_features} and illustrated in Fig.~\ref{fig:face_regions}. This decomposition ensures the representation captures diverse forensic cues rather than relying on a single source.

\begin{table*}[h]
\caption{Complete Feature Set Organized by Forensic Families}
\label{tab:all_features}
\centering
\small
\begin{tabular}{c l c l c l c l}
\toprule
\multicolumn{2}{c}{\textbf{Photometric}} &
\multicolumn{2}{c}{\textbf{Textural}} &
\multicolumn{2}{c}{\textbf{Geometric}} &
\multicolumn{2}{c}{\textbf{Compression}} \\
\cmidrule(r){1-2} \cmidrule(r){3-4} \cmidrule(r){5-6} \cmidrule(r){7-8}
Idx & Variable & Idx & Variable & Idx & Variable & Idx & Variable \\
\midrule
1 & mean\_R & 19 & GLCM\_contrast\_L & 43 & inter\_eye\_dist & 57 & ELA\_mean\_face \\
2 & mean\_G & 20 & GLCM\_contrast\_R & 44 & nose\_to\_mouth\_dist & 58 & ELA\_std\_face \\
3 & mean\_B & 21 & GLCM\_corr\_L & 45 & mouth\_width & 59 & ELA\_mean\_mouth \\
4 & std\_R & 22 & GLCM\_corr\_R & 46 & mouth\_opening & 60 & ELA\_std\_mouth \\
5 & std\_G & 23 & GLCM\_energy\_L & 47 & face\_ratio & 61 & jpeg\_res\_face \\
6 & std\_B & 24 & GLCM\_energy\_R & 48 & jaw\_width & 62 & jpeg\_res\_boundary \\
7 & mean\_H & 25 & GLCM\_homogeneity\_L & 49 & brow\_eye\_dist\_L & 63 & block\_h \\
8 & mean\_S & 26 & GLCM\_homogeneity\_R & 50 & brow\_eye\_dist\_R & 64 & block\_v \\
9 & mean\_V & 27 & entropy\_skin\_L & 51 & eye\_open\_L & 65 & jaw\_contour\_break \\
10 & mean\_Y & 28 & entropy\_skin\_R & 52 & eye\_open\_R & 66 & hair\_contour\_break \\
11 & mean\_Cr & 29 & lap\_var\_L & 53 & mouth\_symmetry & 67 & dct\_hf\_face \\
12 & mean\_Cb & 30 & lap\_var\_R & 54 & nose\_center\_shift & 68 & dct\_hf\_boundary \\
13 & global\_contrast & 31 & hf\_energy\_L & 55 & head\_yaw & & \\
14 & light\_asym\_LR & 32 & hf\_energy\_R & 56 & head\_pitch & & \\
15 & sat\_asym\_LR & 33 & lbp\_mean\_L & & & & \\
16 & ratio\_R\_G & 34 & lbp\_mean\_R & & & & \\
17 & ratio\_B\_G & 35 & grad\_mag\_L & & & & \\
18 & highlight\_prop & 36 & grad\_mag\_R & & & & \\
& & 37 & wavelet\_LH\_L & & & & \\
& & 38 & wavelet\_LH\_R & & & & \\
& & 39 & wavelet\_HL\_L & & & & \\
& & 40 & wavelet\_HL\_R & & & & \\
& & 41 & wavelet\_HH\_L & & & & \\
& & 42 & wavelet\_HH\_R & & & & \\
\bottomrule
\end{tabular}
\end{table*}

\subsection{Photometric Family (18 Variables)}
For each facial region $F$, we explicitly represent color in the RGB, HSV, and YCrCb spaces \cite{Gonzalez2018} and compute the mean of each channel $c \in {R,G,B,H,S,V,Y,Cr,Cb}$ as follows:
\begin{eqnarray}
\mu_c &=& \frac{1}{|F|}\sum_{p\in F} I_p(c)
\end{eqnarray}
In addition to the mean, we also compute the standard deviation for each the RGB channel $c \in \{R,G,B\}$
\begin{eqnarray}
\text{std}\_c &=& \left[\frac{1}{|F|}\sum_{p\in F}\left(I_p(c)-\mu_c\right)^2\right]^{1/2}
\end{eqnarray}
The color ratios are used to capture relative chromatic relationships, and are given as follows:
\begin{equation}
\mathrm{ratio}_{R\_G} = \frac{\mu_R}{\mu_G+\varepsilon},
\qquad
\mathrm{ratio}_{B\_G} = \frac{\mu_B}{\mu_G+\varepsilon}
\end{equation}
To analyze symmetry, the face is partitioned into left and right halves, denoted $I_L$ and $I_R$, respectively. $\varepsilon$ is a small constant that prevents division by 0. Let $\mu_Y^L$ and $\mu_Y^R$ be the mean luminance values computed over the left and right regions, and $\mu_S^L$ and $\mu_S^R$ the corresponding mean saturation values (e.g., in the HSV color space). The illumination and saturation asymmetry measures are then defined as:
\begin{align}
\mathrm{light\_asym} &= \left| \mu_Y^L - \mu_Y^R \right|; ~ 
\mathrm{sat\_asym} = \left| \mu_S^L - \mu_S^R \right|
\end{align}
Here, $Y$ and $S$ denote the luminance and saturation channels in the  YCrCb and HSV color spaces, respectively.

The highlight proportion is defined as
\begin{equation}
\mathrm{highlight\_prop} = \frac{1}{|F|}\sum_{p\in F}\mathbb{I}\left(Y(p)\ge\tau_{\text{high}}\right).
\end{equation}
where $\tau_{\text{high}}$ denotes the luminance threshold above which a pixel is considered a highlight pixel within the facial region.

Finally, although the different color spaces are mathematically related, their explicit inclusion provides complementary chromatic and luminance views, making manipulation-induced inconsistencies more directly accessible and reducing the need for the model to discover such transformations implicitly. These features quantify whether color balance, luminance, saturation, and local illumination remain consistent across the face. As shown in Fig.~\ref{fig:feature_distributions_all}, deepfake manipulations tend to preserve global appearance while introducing subtle per-channel variations and illumination asymmetries.

\subsection{Textural Family (24 Variables)}
To characterize face texture, we first extract statistical features derived from the normalized \textit{gray-level co-occurrence matrix} (GLCM), denoted by $P(i,j)$ where $i$ and $j$ are gray levels, computed for a given inter-pixel displacement $\Delta$. From this matrix, we consider the following standard descriptors: \textit{contrast}, \textit{energy}, \textit{homogeneity}, \textit{correlation}, and \textit{entropy}~\cite{Gonzalez2018}.  

To capture facial asymmetries, all features are computed over the left and right regions of the face. Let $I_L$ and $I_R$ denote the left and right facial regions, respectively, obtained by partitioning the face along a vertical axis passing through its center. For each region, we compute a corresponding GLCM, denoted by $P_L(i,j)$ and $P_R(i,j)$, from which the above statistical descriptors are derived.

Let $\nabla I$ and $\nabla^2 I$ denote the gradient and Laplacian of image $I$. For each facial region, we compute the gradient magnitudes $|\nabla I_L|$ and $|\nabla I_R|$ to measure edge strength, and the Laplacian variances $\mathrm{Var}(\nabla^2 I_L)$ and $\mathrm{Var}(\nabla^2 I_R)$ to capture high-frequency content. In addition, \textit{local binary pattern} (LBP) features are extracted per region, and their mean values serve as compact descriptors of micro-texture~\cite{Ojala94}. Multi-scale texture is further captured via wavelet subband energies (HL, LH, HH) for left and right regions (see Fig. \ref{fig:face_regions}-middle), representing horizontal, vertical, and diagonal components \cite{Allili2012}. Fig.~\ref{fig:feature_distributions_all} shows that fake samples exhibit broader distributions and higher variability, reflecting unstable texture synthesis and imperfect skin detail reproduction \cite{Yapi2023}.

\subsection{Geometric Family (14 Variables)}
To characterize the geometric structure of the face, we extract a set of landmark-based features derived from pairwise distances and simple geometric ratios. Let $\mathbf{l}_a = (x_a, y_a)$ and $\mathbf{l}_b = (x_b, y_b)$ denote two facial landmarks. Their Euclidean distance is defined as
\begin{equation}
d(\mathbf{l}_a,\mathbf{l}_b) = \sqrt{(x_a - x_b)^2 + (y_a - y_b)^2}.
\end{equation}
Such distances provide a direct measure of spatial relationships between key facial components. For instance, the inter-eye distance
\begin{equation}
d_{\text{inter-eye}} = d(\mathbf{l}_{\text{left eye}}, \mathbf{l}_{\text{right eye}})
\end{equation}
captures the face scale, while the nose-to-mouth distance
\begin{equation}
d_{\text{nose-mouth}} = d(\mathbf{l}_{\text{nose tip}}, \mathbf{l}_{\text{mouth center}})
\end{equation}
encodes vertical facial proportions. Similarly, the mouth width
\begin{equation}
w_{\text{mouth}} = d(\mathbf{l}_{\text{mouth left}}, \mathbf{l}_{\text{mouth right}})
\end{equation}
and mouth opening
\begin{equation}
o_{\text{mouth}} = d(\mathbf{l}_{\text{upper lip}}, \mathbf{l}_{\text{lower lip}})
\end{equation}
describe articulation patterns and local deformations associated with speech or expressions.
To capture global facial proportions, we define the face aspect ratio as
\begin{equation}
r_{\text{face}} = \frac{W_{\text{face}}}{H_{\text{face}}},
\end{equation}
where $W_{\text{face}}$ and $H_{\text{face}}$ denote the width and height of the face bounding box, respectively. This ratio provides a coarse descriptor of facial shape and pose variations.

In addition, we quantify bilateral symmetry through a mouth symmetry score
\begin{equation}
S_{\text{mouth}} = \left| d(\mathbf{l}_{\text{left mouth corner}}, \mathbf{l}_c) - d(\mathbf{l}_{\text{right mouth corner}}, \mathbf{l}_c) \right|,
\end{equation}
where $\mathbf{l}_c$ is a reference facial center (e.g., the midpoint between the eyes or the nose tip). This measure reflects deviations from ideal symmetry, which are informative in detecting subtle inconsistencies. These geometric features describe structural consistency, facial proportions, articulation dynamics, symmetry, and coarse pose. While synthetic faces may appear visually convincing, they often exhibit slight geometric inconsistencies or instabilities, particularly across frames. Capturing these variations provides a complementary cue to appearance-based features, enabling more robust analysis of facial authenticity. As illustrated in Fig.~\ref{fig:feature_distributions_all}, geometric features highlight subtle inconsistencies in facial structure and symmetry that are not always perceptible visually.

\subsection{Compression-Related Family (12 Variables)}
Here, we extract a set of features designed to capture inconsistencies introduced by image recompression, blending operations, and block-based coding mechanisms. These features are computed over semantically meaningful regions of the face, such as the full facial area, the mouth region, and boundary zones (e.g., jawline and hair contour), where artifacts are more likely to appear.

Let $\hat{I}_Q$ denote the image recompressed at a given JPEG quality factor $Q$. The error-level analysis (ELA), or residual, is defined as
\begin{equation}
E = |I - \hat{I}_Q|,
\end{equation}
which highlights discrepancies between the original image $I$ and its recompressed version. For a region $R$, we compute the mean $\mu_{\text{ELA}}(R)$ and standard deviation $\sigma_{\text{ELA}}(R)$ of the ELA residual. These statistics capture the average intensity and variability of recompression artifacts, respectively, and are evaluated separately over the face and mouth regions to detect localized inconsistencies. In addition, we compute a JPEG residual energy over region $R$:
\begin{equation}
E_{\text{JPEG}}(R) = \frac{1}{|R|}\sum_{p \in R} r(p)^2,
\end{equation}
where $r(p)$ is  the local recompression residual. This measure reflects the  strength of compression-induced distortions.

\begin{figure*}[p]
    \centering
    \includegraphics[width=\textwidth,height=0.88\textheight,keepaspectratio]{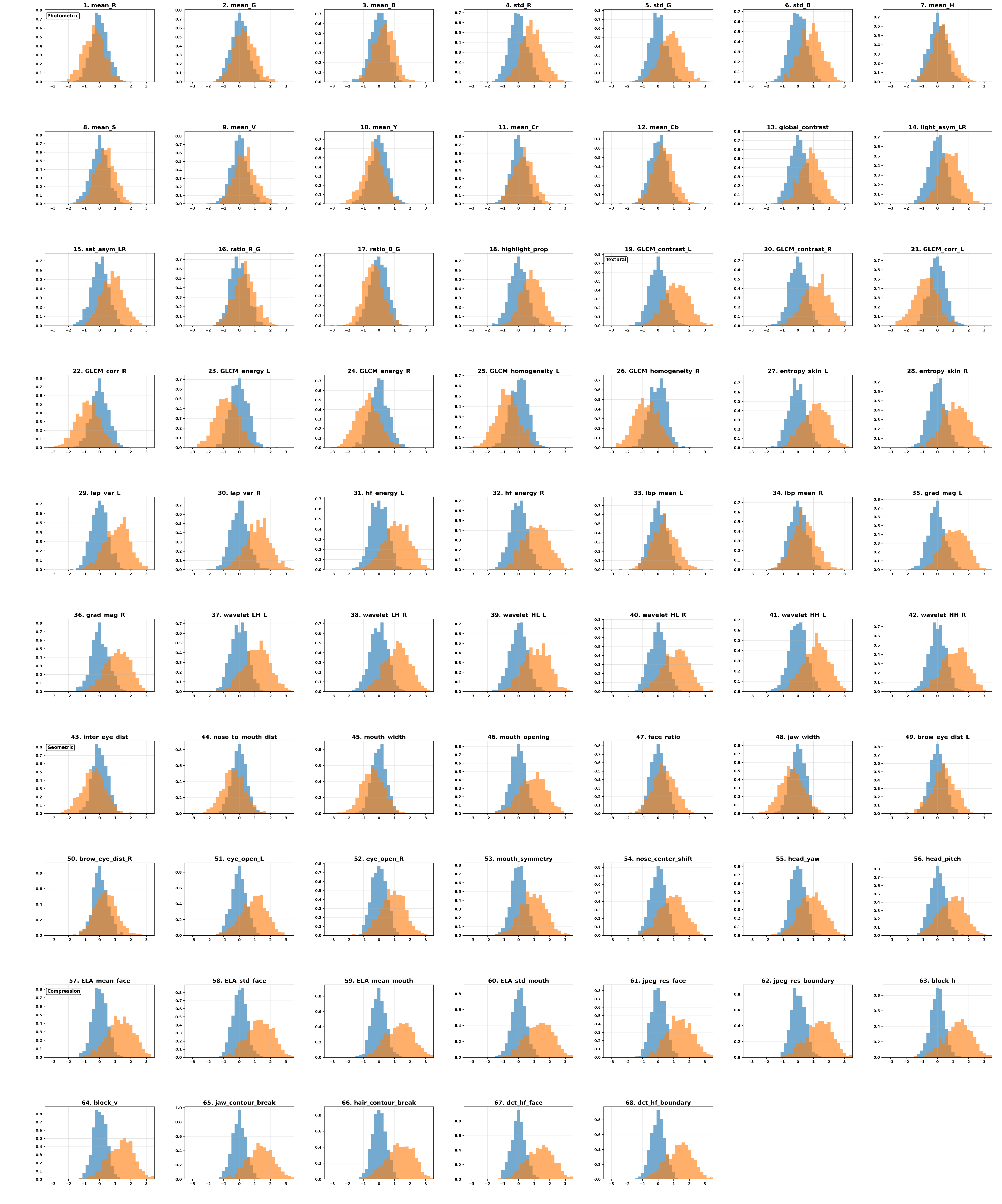}
    \caption{Distribution of selected photometrc, geometric, textural, and compression-related forensic features for real (orange) and fake (blue) samples extracted from the FaceForensic++ dataset.}
    \label{fig:feature_distributions_all}
\end{figure*}

To capture block-based artifacts inherent to JPEG encoding, we define horizontal and vertical blocking scores:
\begin{equation}
B_h = \frac{1}{N_h}\sum_{(p,q)\in\mathcal{B}_h}|I(p)-I(q)|,
\end{equation}
\begin{equation}
B_v = \frac{1}{N_v}\sum_{(p,q)\in\mathcal{B}_v}|I(p)-I(q)|.
\end{equation}
where $\mathcal{B}_h$ and $\mathcal{B}_v$ denote sets of pixel pairs located across horizontal and vertical block boundaries, respectively. These measures quantify discontinuities introduced by block-based compression. Furthermore, contour-based features such as \textit{jaw\_contour\_break} and \textit{hair\_contour\_break} are designed to detect structural discontinuities along facial boundaries, which may arise from imperfect blending or compositing. Finally, high-frequency components are analyzed through the energy of discrete cosine transform (DCT) coefficients in selected regions (face and boundary), capturing fine-scale compression artifacts that are often altered in synthetic content. Collectively, these features identify traces left by manipulation processes, blending, recompression, and block-structured coding artifacts, which tend to concentrate around regions of high motion or semantic importance such as the mouth, jawline, and facial boundaries.
As shown in Fig.~\ref{fig:feature_distributions_all}, feature distributions differ significantly between real and fake samples, confirming that each family captures complementary forensic evidence, with compression-based features providing particularly strong discriminative signals even in visually high-quality deepfakes.

\subsection{LSTM-Based Deepfake Prediction}

Given the structured temporal representation $\mathbf{X}\in\mathbb{R}^{20\times68}$, the prediction task consists in modeling the temporal evolution of explicit forensic variables to identify inconsistencies indicative of deepfake generation. Unlike frame-based approaches, the objective here is not only to detect anomalous values, but to capture their dynamics over time, such as instability, drift, or incoherent co-evolution across feature families.

To this end, we employ a Long Short-Term Memory (LSTM) network, which is well-suited for sequential data with temporal dependencies. The LSTM processes the sequence frame by frame, allowing the model to learn both short-term variations and longer-range temporal patterns within the tracked face trajectory. This is particularly relevant in our setting, where deepfake artifacts often manifest as subtle temporal inconsistencies rather than strong spatial distortions.

Each frame is represented by a structured feature vector

\begin{equation}
\mathbf{x}_t = \left[\mathbf{x}^{(p)}_t,\mathbf{x}^{(tx)}_t,\mathbf{x}^{(g)}_t,\mathbf{x}^{(c)}_t\right]\in\mathbb{R}^{68},
\end{equation}

where the components correspond to 18 photometric, 24 textural, 14 geometric, and 12 compression-related descriptors. A temporal sample is constructed by stacking 20 consecutive frames from the same tracked face, yielding a matrix $\mathbf{X}\in\mathbb{R}^{20\times68}$, where rows encode temporal evolution and columns correspond to explicit forensic variables. This representation preserves both temporal coherence and semantic structure, enabling the analysis of feature dynamics across time.

The classification model operates directly on this structured sequence. The architecture consists of a two-stage temporal encoder followed by a compact decision head:
\begin{small}
\begin{equation}
\mathbf{X} \rightarrow \text{LSTM}(128) \rightarrow \text{LSTM}(64) \rightarrow \text{Dense}(32) \rightarrow \text{Sigmoid}
\end{equation}
\end{small}
The first LSTM layer (128 units) returns full sequences, allowing the second LSTM layer (64 units) to capture higher-level temporal dependencies. The second LSTM layer returns the final hidden state, which is then passed to the fully connected layers. Both LSTM layers are unidirectional and include dropout regularization (rate = 0.2) to improve generalization and reduce overfitting.

This design captures both short-term fluctuations and long-range dependencies within the sequence, while preserving interactions across feature families. By modeling the temporal evolution of explicit variables rather than latent embeddings, the network focuses on interpretable forensic patterns such as instability, drift, and cross-feature inconsistencies. The model is trained using the Adam optimizer with a learning rate of $10^{-3}$, a batch size of 32, and a maximum of 35 epochs, with early stopping (patience of 5) based on validation loss. The final classification uses a sigmoid activation optimized with binary cross-entropy loss. The full pipeline runs in real time, processing each video frame within the latency budget required for live deployment.

\section{Experimental results}
\label{experiments}

\subsection{Dataset-Wise Corpus Construction and Standardization}

The protocol uses four widely adopted deepfake video datasets: FaceForensics++ \cite{Rossler2019FaceForensics}, Celeb-DF v2 \cite{Li2020CelebDF}, a subset of the DeepFake Detection Challenge dataset (DFDC) \cite{Dolhansky2020DFDC}, and DeeperForensics \cite{Jiang2020DeeperForensics}. Unlike a merged-corpus protocol, the proposed methodology is applied independently to each dataset. Each dataset therefore undergoes the same preprocessing, tracking, sequence construction, feature extraction, and LSTM-based classification pipeline, but the reported results are dataset-specific.

Having multiple datasets reflects different manipulation pipelines, acquisition conditions, compression levels, and visual realism. It also avoids hiding dataset-specific difficulty behind a single aggregate score. The method is therefore evaluated not only by its strongest result, but also by its behavior across heterogeneous data sources. The fixed parameters of the protocol are:
\begin{itemize}[leftmargin=1.4em]
    \item Harmonized frame rate: 25 fps after resampling;
    \item Final face crop size: $224 \times 224$ pixels;
    \item Sequence length: 20 consecutive frames. One sequence corresponds to one tracked face only
    \item Sequence stride: 10 frames;  
\end{itemize}

For each dataset, preprocessing converts raw videos into standardized face trajectories by discarding low-quality detections, resizing valid face crops to $224\times224$, and generating fixed-length temporal windows. Each retained sequence is encoded as a $20\times68$ matrix, and the train/validation/test split is performed at the video level to prevent temporal leakage. This pipeline transforms heterogeneous inputs, varying in frame rate, resolution, number of faces, and quality, into consistent temporal samples of a single tracked identity. Such standardization ensures that the temporal model receives inputs with fixed sequence length, uniform spatial resolution, consistent feature dimensionality, preserved temporal order, and identity continuity, thereby avoiding unstable or semantically inconsistent representations.

To evaluate the proposed methods, we report standard metrics including accuracy, precision, recall, and F1-score, capturing both overall performance and class-wise behavior. Accuracy reflects global correctness, while precision and recall measure the reliability of positive predictions and the ability to detect all relevant instances, particularly important in imbalanced settings. The F1-score balances this trade-off. We also report the Area Under the ROC Curve (AUC), which assesses discriminative ability across all decision thresholds and provides insight into robustness under varying conditions.

\subsection{Dataset-wise Main Results}

Table~\ref{tab:dataset_results} reports the performance obtained when the same proposed pipeline is applied independently to each dataset. Unlike the global performance reported in the ablation study (97.8\% accuracy), which is computed over the complete test set, the results presented here are evaluated separately on each dataset and therefore reflect dataset-specific behavior rather than a global aggregate.

The highest performance is observed on FaceForensics++, where the method reaches 98.0\% accuracy and 98.0\% F1-score, indicating that the proposed explicit feature representation effectively captures compression inconsistencies, photometric imbalances, and fine-grained textural artifacts present in this benchmark. Similarly, the DFDC subset achieves strong performance (97.6\% accuracy and 97.6\% F1-score), suggesting that the learned temporal dynamics and explicit forensic descriptors generalize well to moderately realistic conditions.

DeeperForensics also maintains high performance (96.2\% accuracy and 96.2\% F1-score), demonstrating robustness to controlled perturbations. In contrast, Celeb-DF v2 remains the most challenging dataset, with 91.0\% accuracy and 91.0\% F1-score. This behavior is consistent with its higher visual realism, where facial manipulations exhibit fewer low-level inconsistencies, thereby reducing the separability of photometric, textural, geometric, and compression-related descriptors. These results highlight that the proposed model performs consistently across datasets while reflecting the intrinsic difficulty associated with increasingly realistic deepfake generation. However, this performance gap across datasets indicates that the discriminative power of explicit feature-based representations remains dependent on the visibility of low-level artifacts, which progressively diminish in more advanced and visually realistic deepfake generation pipelines.

\begin{table}[H]
\caption{Dataset-wise performance of the proposed method. The last row reports the unweighted arithmetic average across datasets, not the global test-set performance.}
\label{tab:dataset_results}
\centering
\begin{tabular}{lccccc}
\toprule
\textbf{Dataset} & \textbf{Acc.} & \textbf{Prec.} & \textbf{Rec.} & \textbf{F1} & \textbf{AUC} \\
\midrule
FaceForensics++ & 98.0\% & 97.9\% & 98.1\% & 98.0\% & 0.995 \\
Celeb-DF v2 & 91.0\% & 90.8\% & 91.2\% & 91.0\% & 0.948 \\
DFDC subset & 97.6\% & 97.4\% & 97.8\% & 97.6\% & 0.991 \\
DeeperForensics & 96.2\% & 96.0\% & 96.4\% & 96.2\% & 0.984 \\
\midrule
Average & 95.7\% & 95.5\% & 95.9\% & 95.7\% & 0.980 \\
\bottomrule
\end{tabular}
\end{table}

The unweighted dataset average assigns equal importance to each dataset regardless of its number of test sequences. In contrast, the global accuracy reported in the ablation study is computed over the complete test set and is therefore sequence-level and dependent on dataset size distribution.

Table~\ref{tab:cm} illustrates the behavior of the classifier over a balanced test subset. This confusion matrix is computed on a balanced illustrative subset and may slightly differ from the overall dataset-level performance due to sampling effects. It shows that 982 real sequences were correctly recognized, 18 real sequences were misclassified as fake, 24 fake sequences were missed, and 976 fake sequences were correctly detected. This confirms that the system detects the large majority of fake sequences while maintaining a low false-positive rate on real sequences.

\begin{table}[H]
\caption{Illustrative Confusion Matrix on a Balanced Test Subset}
\label{tab:cm}
\centering
\begin{tabular}{|l|c|c|}
\hline
 & \textbf{Predicted Real} & \textbf{Predicted Fake} \\
 \hline
\textbf{Real} & 982 & 18 \\
\hline
\textbf{Fake} & 24 & 976 \\
\hline
\end{tabular}
\end{table}

\subsection{Comparison with Recent Literature}

\begin{table*}[t]
\caption{Quantitative comparison with recent deepfake detection methods. The table reports only the metrics explicitly provided by the corresponding papers. Since protocols, datasets, and evaluation settings differ across works, this table is intended as a quantitative reference comparison rather than a strict leaderboard.}
\label{tab:quantitative_comparison}
\centering
\small
\begin{tabular}{|
>{\raggedright\arraybackslash}p{2.0cm}|
>{\raggedright\arraybackslash}p{0.8cm}|
>{\raggedright\arraybackslash}p{6 cm}|
>{\raggedright\arraybackslash}p{5.4cm}|
p{1.5cm}|}

\hline
\textbf{Method} & \textbf{Year} & \textbf{Method type / Protocol} & \textbf{Reported quantitative results} & \textbf{Avg. AUC} \\
\hline

Amin et al. \cite{Amin2024TemporalCoherence} 
& 2024 
& Temporal facial pattern and coherence analysis; evaluated on FF++, DFD, DFDC, DeeperForensics 
& AUC: 98.82\% (FF++); 91.09\% (DFD); 86.94\% (DFDC); 98.80\% (DFo) 
& 93.91\% \\
\hline

Yan et al. \cite{Yan2025PlugPlay} 
& 2025 
& Video-level blending and spatiotemporal tuning; train FF++ c23, test CDF-v2 / DFDC 
& AUC: 94.7\% (CDF-v2); 84.3\% (DFDC) 
& 89.5\% \\
\hline

Kim et al. \cite{Kim2025TemporalFrequency} 
& 2025 
& Pixel-wise temporal-frequency modeling; cross-dataset evaluation 
& Video AUC: 89.7\% (CDF); 75.2\% (DFDC); 99.3\% (FSh); 99.4\% (DFo); 97.3\% (DFD) 
& 92.2\% \\
\hline

Guo et al. \cite{Guo2025TemporalForgery} 
& 2025 
& Temporal forgery cue modeling; cross-dataset Celeb-DF / DFDC / FFIW 
& Video AUC: 93.18\% (Celeb-DF); 86.05\% (DFDC); 91.27\% (FFIW) \newline
Frame AUC: 91.38\% (Celeb-DF); 85.43\% (DFDC); 90.21\% (FFIW)
& 90.17\% \\
\hline

Hu et al.~\cite{Hu2025HICOM} & 2025 &
Multi-face detection; human-inspired scene-motion, inter-face appearance, gaze, and face-body cues; evaluated on FFIW, OpenForensics, and DF-Platter &
FAU/FCAU: 95.9\%/92.1\% (FFIW); 99.9\%/98.9\% (OpenForensics); 98.4\%/94.6\% (DF-Platter) &
95.2\% \\
\hline

\textbf{Our method} 
& 2026 
& Multi-dataset evaluation on FaceForensics++, Celeb-DF v2, DFDC subset, and DeeperForensics using tracked-face temporal windows and LSTM modeling 
& Accuracy: \textbf{97.8\%}; Precision: \textbf{97.6\%}; Recall: \textbf{98.0\%}; F1-score: \textbf{97.8\%}; pooled AUC: \textbf{99.2\%} 
& \textbf{98.0\%} \\
\hline

\end{tabular}
\end{table*}

Table~\ref{tab:quantitative_comparison} shows a comparison of our approach with regard to recent state-of-the art methods. Since not all works report the same evaluation metrics, the comparison is limited to AUC  values explicitly provided in the original papers. The comparative methods are: 1) Amin et al. \cite{Amin2024TemporalCoherence}  which focus on temporal consistency patterns across frames, 2) Yan et al. \cite{Yan2025PlugPlay} uses spatiotemporal adaptation mechanisms to improve cross-dataset generalization, 3)Kim et al. \cite{Kim2025TemporalFrequency}  exploits temporal-frequency inconsistencies at the pixel level, capturing subtle artifacts that are not observable in the spatial domain alone.  Guo et al. \cite{Guo2025TemporalForgery} extended this direction by explicitly modeling temporal forgery cues, reporting both frame-wise and video-wise performance, 5) Hu et al. \cite{Hu2025HICOM} which incorporate contextual information in multi-face scenarios (e.g. gaze alignment, inter-person interactions, and motion coherence). 

Despite differences in evaluation settings, several trends emerge. Temporal coherence and spatiotemporal adaptation methods \cite{Amin2024TemporalCoherence,Yan2025PlugPlay} perform strongly on controlled datasets but degrade on challenging benchmarks such as DFDC, indicating limited robustness to domain shifts. Frequency- and temporal-cue approaches \cite{Kim2025TemporalFrequency,Guo2025TemporalForgery} achieve competitive results but vary across datasets, while context-aware methods \cite{Hu2025HICOM} improve robustness in multi-face scenarios. In contrast, the proposed method achieves the highest average AUC (98.0\%) and a strong pooled AUC (99.2\%), with consistently high precision, recall, and F1-score. These results suggest that combining explicit forensic features with temporal modeling provides a complementary representation of deepfake artifacts and improves consistency across diverse datasets.

\subsection{Cross-Dataset Generalization}

To further assess robustness and generalization, we evaluate the proposed framework on two unseen datasets: ForgeryNet \cite{he2021forgerynet} as a validation set and WildDeepfake \cite{zi2020wilddeepfake} as a held-out test set. The same pipeline and model weights learned from the original four training datasets are applied without retraining or fine-tuning, providing a strictly zero-shot cross-dataset evaluation.

ForgeryNet contains diverse manipulation types under moderate and relatively controlled compression conditions, making it suitable for evaluating generalization to unseen forgery pipelines. In contrast, WildDeepfake comprises real-world deepfake videos collected from the Internet, with substantial variability in acquisition conditions, compression levels, video quality, and generation methods. As reported in Table~\ref{tab:cross_results}, the model achieves 93.8\% accuracy and F1-score on ForgeryNet, and 89.4\% accuracy and 89.5\% F1-score on WildDeepfake, respectively. These results demonstrate strong cross-dataset generalization without retraining or adaptation, indicating that the learned temporal patterns are not limited to the distributions of the training datasets. The lower performance on WildDeepfake reflects the greater difficulty of in-the-wild videos, where uncontrolled acquisition, repeated post-processing, and compression variability can weaken or obscure low-level forensic cues. Nevertheless, the sustained performance suggests that combining complementary forensic descriptors with temporal modeling provides useful robustness under substantial distribution shifts.

\begin{table}[t]
\caption{Cross-dataset generalization results. ForgeryNet is used
as validation and WildDeepfake as test. No retraining is performed.}
\label{tab:cross_results}
\centering
\begin{tabular}{lccccc}
\toprule
Dataset &  Acc. & Prec. & Rec. & F1 & AUC \\
\midrule
ForgeryNet    &  93.8\% & 93.5\% & 94.1\% & 93.8\% & 0.967 \\
WildDeepfake  &  89.4\% & 88.8\% & 90.2\% & 89.5\% & 0.941 \\
\bottomrule
\end{tabular}
\end{table}

Table~\ref{tab:cross_confusion} presents the illustrative confusion matrices
for each cross-dataset evaluation. On ForgeryNet, 935 real
sequences are correctly classified while 65 are misclassified
as fake, and 941 fake sequences are correctly detected while
59 are missed. On WildDeepfake, 887 real sequences are
correctly classified with 113 false positives, and 902 fake
sequences are correctly detected with 98 missed. These results
confirm that the model maintains a balanced detection behavior,
with no systematic bias toward either class.

\begin{table}[t]
\caption{Illustrative confusion matrices for cross-dataset evaluation.}
\label{tab:cross_confusion}
\centering
\begin{tabular}{llcc}
\toprule
Dataset & Ground Truth & Pred. Real & Pred. Fake \\
\midrule
\multirow{2}{*}{ForgeryNet}   & Real & 935 & 65  \\
                               & Fake & 59  & 941 \\
\midrule
\multirow{2}{*}{WildDeepfake} & Real & 887 & 113 \\
                               & Fake & 98  & 902 \\
\bottomrule
\end{tabular}
\end{table}

\subsection{Ablation Study}
Tables~\ref{tab:cross_ablation_forgery} and~\ref{tab:cross_ablation_wild} report the feature-family ablation results on ForgeryNet and WildDeepfake, respectively. Across both datasets, the best performance is consistently achieved when all four feature families are jointly used, while removing any single family leads to a systematic degradation across all metrics. This confirms that the proposed representation benefits from the integration of heterogeneous forensic evidence rather than relying on a dominant individual cue. Moreover, the same importance ranking is observed on both datasets: \textit{compression} $>$ \textit{photometric} $>$ \textit{textural} $>$ \textit{geometric}. This trend  suggests that the relative contribution and complementarity of the feature families persist across different data distributions rather than reflecting overfitting to a particular benchmark.

Compression and photometric features have the largest impact, capturing persistent traces such as recompression artifacts, color and illumination inconsistencies, and blending irregularities. Textural and geometric features contribute complementary cues related to skin micro-patterns and facial structure, particularly when dominant artifacts are weakened by post-processing. Performance drops are systematically larger on WildDeepfake than on ForgeryNet; for instance, removing compression features reduces the F1-score by $3.9$ and $3.2$ points, respectively. This suggests that challenging in-the-wild conditions reduce the reliability of individual cues, increasing the importance of combining heterogeneous evidence. Overall, the results confirm that the framework's strength lies in integrating complementary forensic features whose temporal evolution reveals subtle manipulation traces that may remain ambiguous within a single domain or frame.

\begin{table}[t]
\caption{Ablation on feature families --- ForgeryNet (Validation).}
\label{tab:cross_ablation_forgery}
\centering
\begin{tabular}{lccccc}
\toprule
Configuration & Acc. & Prec. & Rec. & F1 & AUC \\
\midrule
All 4 families      & 93.8\% & 93.5\% & 94.1\% & 93.8\% & 0.967 \\
Without photometric & 90.9\% & 90.5\% & 91.3\% & 90.9\% & 0.941 \\
Without textural    & 91.6\% & 91.2\% & 92.0\% & 91.6\% & 0.948 \\
Without geometric   & 92.0\% & 91.7\% & 92.4\% & 92.0\% & 0.953 \\
Without compression & 90.6\% & 90.1\% & 91.1\% & 90.6\% & 0.938 \\
\bottomrule
\end{tabular}
\end{table}

\begin{table}[t]
\caption{Ablation on feature families --- WildDeepfake (Test).}
\label{tab:cross_ablation_wild}
\centering
\begin{tabular}{lccccc}
\toprule
Configuration & Acc. & Prec. & Rec. & F1 & AUC \\
\midrule
All 4 families      & 89.4\% & 88.8\% & 90.2\% & 89.5\% & 0.941 \\
Without photometric & 85.8\% & 85.0\% & 86.9\% & 85.9\% & 0.902 \\
Without textural    & 86.7\% & 86.0\% & 87.6\% & 86.8\% & 0.913 \\
Without geometric   & 87.2\% & 86.6\% & 88.0\% & 87.3\% & 0.919 \\
Without compression & 85.4\% & 84.6\% & 86.6\% & 85.6\% & 0.899 \\
\bottomrule
\end{tabular}
\end{table}

\section{Conclusion and discussion}
\label{sec:conclusion}
The proposed framework relies on explicitly designed forensic features (photometric consistency, texture statistics, geometric structure, and compression artifacts), which are interpretable, computationally lightweight, and physically intuitive, making their discriminative behavior directly explainable without requiring any post-hoc explanation. Integrated within a temporally coherent tracking pipeline, these features provide a transparent and efficient alternative to black-box deep representations. Ensuring each sequence corresponds to a single identity allows feature variations to reflect meaningful temporal patterns, while LSTM-based aggregation accumulates weak frame-level cues into a robust, interpretable representation. Strong results on several datasets confirm  that representing each sequence as a matrix of explicit forensic features enables the model to capture temporal inconsistencies while preserving interpretability. The approach  demonstrates that physically meaningful features, consistently tracked over time, provide an effective basis for deepfake detection. Cross-dataset generalization experiments on ForgeryNet and WildDeepfake further confirm that the explicit forensic representation transfers to unseen data without retraining, yielding competitive results under challenging in-the-wild conditions.

Looking ahead, several directions merit exploration. First, extending the framework to analyze faces across arbitrary orientations, scales, and pose variations could improve robustness beyond predominantly frontal views, while modeling facial expression dynamics may reveal unnatural affective transitions or temporal inconsistencies. Second, reformulating the facial landmark graph as an explicit graph neural network \cite{Attoumi2025} would allow relational reasoning over facial topology, enabling the model to capture structural anomalies and asymmetries that sequential models cannot directly exploit. Third, incorporating multimodal signals by pairing facial features with speech~\cite{Temmar2025}, phoneme-visual alignment, or physiological cues  could expose cross-modal inconsistencies that unimodal analysis misses. Fourth, replacing the LSTM with a Transformer-based sequence model would enable direct long-range attention over facial trajectories, capturing subtle non-local temporal anomalies beyond the reach of recurrent architectures. Finally, hybrid architectures combining the interpretable handcrafted features introduced here with learned deep representations could improve robustness to unseen manipulations while preserving the transparency and physical grounding required for forensically credible analysis.

\section{Acknowledgement}
\label{acknowledgement}
This work has been achieved thanks to the  the support of the Natural Sciences and Engineering Research Council (RGPIN-2020-05653) and the Centre interdisciplinaire de recherche et d'innovation
en cybersecurité  et société (CIRICS) through the Fonds de recherche du Québec.

\end{document}